\documentclass[letterpaper, 10 pt, journal, twoside]{IEEEtran}  

\IEEEoverridecommandlockouts                              

\usepackage{times}

\usepackage[utf8]{inputenc} 
\usepackage[T1]{fontenc}    
\usepackage{url}            
\usepackage{booktabs}       
\usepackage{amsfonts}       
\usepackage{nicefrac}       
\usepackage{microtype}      
\usepackage{balance}      
\usepackage{wrapfig}
\usepackage[bookmarks=true]{hyperref}
\usepackage{color, colortbl, xcolor}

\title{\LARGE \bf
Visual Navigation Among Humans With Optimal Control as a Supervisor
}

\author{Varun Tolani$^{1*}$, Somil Bansal$^{2*}$, Aleksandra Faust$^{3}$, and Claire Tomlin$^{1}$
\thanks{This research is supported in part by the DARPA Assured Autonomy program under agreement number FA8750-18-C-0101, by NSF under the CPS Frontier project VeHICaL project (1545126), by SRC under the CONIX Center, and by the Google-BAIR Commons program.} 

\thanks{*Equal contribution.}
\thanks{$^{1}$Authors are with EECS at UC Berkeley: \{vtolani, tomlin\}@berkeley.edu. $^{2}$Author is with ECE at University of Southern California: \{somilban\}@usc.edu. $^{3}$Author is with Google Research: \{faust\}@google.com.}

}

\usepackage{graphicx,caption,subcaption,xspace}
\usepackage{amsmath,amssymb,stmaryrd,mathtools}
\usepackage{algorithm}
\usepackage{algpseudocode}
\usepackage{acronym}
\usepackage{comment}
\usepackage{color}
\usepackage{units}
\usepackage{textcomp}
\usepackage{wrapfig}

\usepackage[capitalise]{cleveref}

\DeclarePairedDelimiter{\norm}{\lVert}{\rVert}
\DeclareCaptionLabelSeparator{periodspace}{.\quad}
\crefname{equation}{}{} 
\crefname{section}{Sec.}{Sec.}

\newcommand{\state}{z^V} 
\newcommand{\ctrl}{u^V} 
\newcommand{\pos}{p} 

\renewcommand{\time}{t}

\newcommand{\newdataset}{HumANav\xspace}

\newcommand{\stateH}{z^H}
\newcommand{\ctrlH}{u^H} 

\newcommand{\metNameFov}{LB-WayPtNav-DH-FOV\xspace}
\newcommand{\metNameCorl}{LB-WayPtNav\xspace}

\newcommand{\metName}{LB-WayPtNav-DH\xspace}

\newtheorem{remark}{Remark}

\newcommand{\SBnote}{\textcolor{black}}

\algdef{SE}[SUBALG]{Indent}{EndIndent}{}{\algorithmicend\ }%
\algtext*{Indent}
\algtext*{EndIndent}

\begin{document}

\maketitle


\begin{abstract}
    Real world visual navigation requires robots to operate in unfamiliar, human-occupied dynamic environments.
    Navigation around humans is especially difficult because it requires anticipating their future motion, which can be quite challenging.
    We propose an approach that combines learning-based perception with model-based optimal control to navigate among humans based only on monocular, first-person RGB images.
    Our approach is enabled by our novel data-generation tool, \newdataset, that allows for photorealistic renderings of indoor environment scenes with humans in them, which are then used to train the perception module \textit{entirely} in simulation. 
    Through simulations and experiments on a mobile robot, we demonstrate that the learned navigation policies can anticipate and react to humans without explicitly predicting future human motion,
    generalize to previously unseen environments and human behaviors, and transfer directly from simulation to reality.
    Videos describing our approach and experiments, as well as a demo of \newdataset are available on the project website\footnote{Project website: \href{https://smlbansal.github.io/LB-WayPtNav-DH/}{\textcolor{blue}{https://smlbansal.github.io/LB-WayPtNav-DH/}}}.
\end{abstract}


\section{Introduction} \label{sec:intro}
\IEEEPARstart{A}{utonomous} robot navigation has the potential to enable many critical robot applications, from service robots that deliver food and medicine, to logistics and search and rescue missions. In all these applications, it is imperative for robots to work safely among humans and be able to adjust their own motion plans based on observed human behavior.

One way to approach the problem of autonomous robot navigation among people is to identify humans in the scene, predict their future motion, and react to them safely. 
However, human recognition can be difficult because people come in different shapes and sizes, and might even be partially occluded. 
Human motion prediction, on the other hand, is challenging because the human's navigational goal (intent) is not known to the robot, and people have different temperaments and physical abilities which affect their motion (speed, paths etc.)~\cite{rudenko2019human, zeng2019end}.
These aspects make navigating around humans particularly challenging, especially when the robot itself is operating in a new, \textit{a priori} unknown environment.
Alternative approaches employ end-to-end learning to sidestep explicit recognition and prediction
\cite{unstuck-dinesh,local-replan}.
These methods, however, tend to be sample inefficient and overspecialized to the system on which they were trained~\cite{bansal2019combining, kaufmann2018deep}.

\begin{figure*}
  \centering
    \includegraphics[width=.62\textwidth]{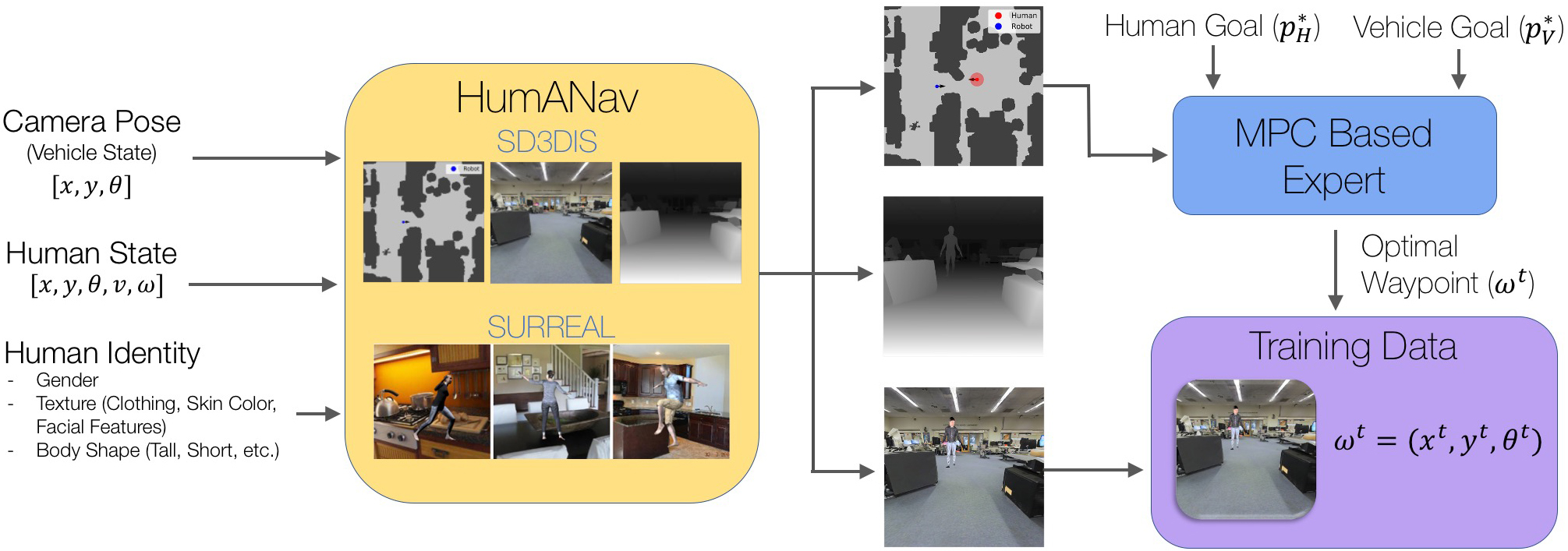}
	\includegraphics[width=.25\textwidth]{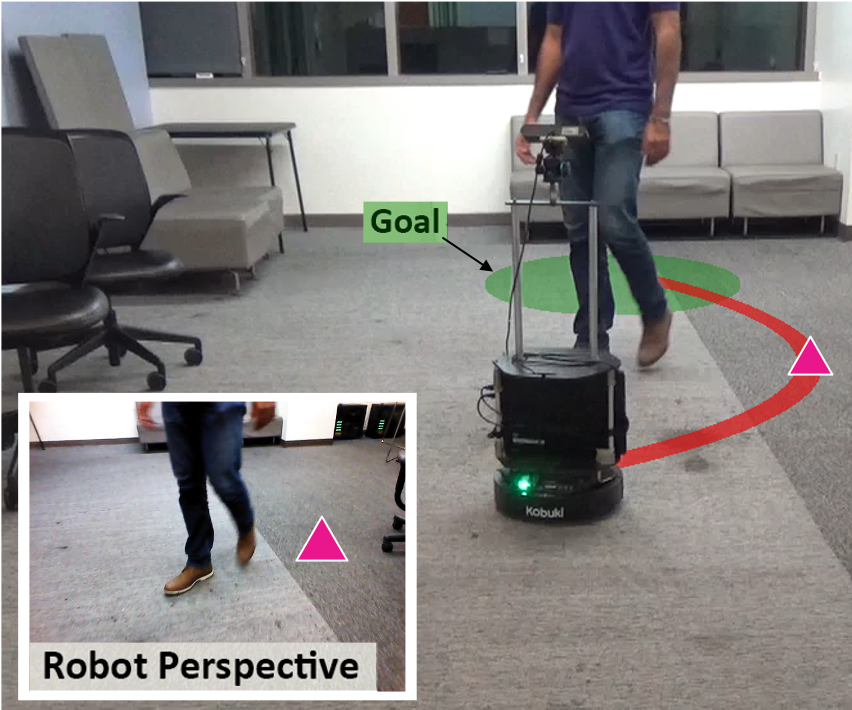}
	\caption{We consider the problem of autonomous visual navigation in \textit{a priori} unknown, indoor environments \emph{with} humans. 
	Our approach, \metName, consists of a learning-based perception module and a model-based planning and control module. To learn navigational behavior around humans, we create the \newdataset data-generation tool which allows for photorealistic renderings in simulated buildings environments \emph{with} humans (left). We use an MPC-based expert along with \newdataset to train \metName \textit{entirely} in simulation. At test time, \metName navigates efficiently in never-before-seen buildings based only on monocular RGB images \emph{and} demonstrates zero-shot, sim-to-real transfer to novel, real buildings around real humans (right).}
    \vspace{-2em}
    \label{fig:front_figure}
\end{figure*}

In this work, we propose an approach to visual navigation among humans based only on monocular RGB images received from an onboard camera, as shown in Fig. \ref{fig:front_figure}.
Our approach is enabled by a novel data-generation tool which we have designed, the Human Active Navigation Data-Generation Tool (\newdataset), a photo-realistic rendering engine for images of humans moving in indoor environments.
Equipped with this data generation tool, we train a modular architecture that combines a learning-based perception module with a dynamics model-based planning and control module to learn navigation policies \textit{entirely} in simulation. 
The photorealistic nature of \newdataset allows for zero-shot, simulation-to-reality transfer of the learned navigation policies, without requiring any expensive demonstrations by an expert or causing any privacy and logistical challenges associated with human subjects.

Our navigation pipeline leverages a learned Convolutional Neural Network (CNN) to predict a waypoint, or the vehicle's next desired state, using the observed RGB image, and uses optimal control to actually reach the waypoint. However, generating supervision for training the CNN in dynamic environments in challenging as (a) it requires simulation of visually realistic humans and their motion, and (b)
the robot motion affects the future scenes so the dataset needs to be \textit{active} (or on-policy) to enable rich human-robot interactions. 
 
To address the above challenges, we propose \newdataset which consists of scans of 6000 synthetic but realistic humans from the SURREAL dataset \cite{varol17_surreal} placed in office buildings from Stanford Large Scale 3D Indoor Spaces Dataset (SD3DIS) \cite{armeni_cvpr16}. \newdataset allows for user manipulation of human agents within the building and provides photorealistic renderings of the scene (RGB, Depth, Surface Normals, etc.). Critically, \newdataset also ensures that important visual cues associated with human movement are present in images (\SBnote{e.g., the legs of a very slow moving human will be closer together compared to those of a very fast moving human}), facilitating reasoning about human motion.
To train the CNN, we propose a fully automated, self-supervision method that uses Model Predictive Control (MPC) along with \newdataset to generate rendered RGB images and corresponding optimal waypoints.

To summarize, the key contributions of this paper are: a)
    \textit{\newdataset}, an active data-generation tool to benchmark visual navigation algorithms around humans; 
    b) \textit{a fully automated self-supervised training scheme via MPC} that leverages \newdataset to generate data for learning navigation policies without requiring any expensive demonstrations by an expert; and
    c) \textit{an autonomous visual navigation method} that uses this training data to learn to navigate around humans in unknown indoor environments based only on monocular RGB images, does not require explicit state estimation and trajectory prediction of the human, and performs zero-shot transfer of learned policies from simulation to reality.
\section{Related Work} \label{sec:related_work}
\textbf{Visual Navigation:} 
An extensive body of research studies autonomous visual navigation in static environments using RGB images~\cite{zhu2016target, gupta2017cognitive, khan2017memory, muller2018driving, meng2019neural, pan2017agile, savva2019habitat},
laser scans~\cite{tai2017virtual}, and top views~\cite{bansal2018chauffeurnet}.
Other works have looked into visual locomotion~\cite{gandhi2017learning, kahn2017uncertainty, sadeghi2016cadrl, loquercio2018dronet}, servoing~\cite{hirose2019deep, sadeghi2019divis}, aggressive driving~\cite{jung2018perception, pmlr-v78-drews17a, kaufmann2018deep}, and topological memory for navigation \cite{chen2019behavioral, savinov2018semi, singh2020neural}.
\cite{bansal2019combining} uses a learning-based perception module to output waypoints for a model-based planner to navigate in novel static environments. 
Our navigation pipeline uses a similar decomposition between perception and planning, but the waypoints are learned to additionally anticipate and react to future human motion.
\SBnote{To learn navigation policies, our framework imitates an expert with privileged information. 
This approach has shown promising results for indoor navigation \cite{bansal2019combining}, autonomous driving \cite{chen2020learning}, and drone acrobatics \cite{kaufmann2020deep}.} 

Other works \cite{local-replan,prm-rl} use classical planning in static environments as the higher level planner, along with reinforcement learning for adaptive local planning and path tracking in dynamic environments, or train in photorealistic static environments \cite{gao2017intention} and evaluate in dynamic environments.
This approach limits their ability to reason about the dynamic nature of human and planning optimal paths around it. 
Instead, we learn the waypoint placement for the high-level planner to be optimal with respect to the human motion, and use optimal control to reach the waypoint.

\textbf{Navigation among humans:}
Classical robotics factorizes the problem of navigation among humans into sub-problems of detection and tracking \cite{brunetti2018computer}, human motion prediction \cite{rudenko2019human}, and planning \cite{lavalle2006planning}.
However, reliable state estimation of the human might be challenging, especially when the robot is using narrow field-of-view sensors such as a monocular RGB camera.
Moreover, human motion prediction itself can be quite challenging and is an active area of research \cite{rudenko2019human}. 
Learning-based approaches have also been explored to produce socially-compliant motion among humans~\cite{li2018role}; however, the method requires human trajectories, and relies on detection and tracking algorithms to locate the humans.
Other methods use depth sensors \cite{pfeiffer2017perception,unstuck-dinesh,chen2019crowd, sathyamoorthy2020densecavoid, everett2018motion} to navigate in crowded spaces. 
These methods do not require high visual fidelity, but require expensive wide-field of view LiDAR sensors.
Our method predicts goal-driven waypoints to navigate around humans using \textit{only} a monocular RGB image, without explicitly estimating human state or motion. 

\textbf{Social visual navigation datasets:}
\cite{martin2019jrdb} proposes a dataset on multi-modal social visual navigation, collected in real environments using real humans, manual annotation, and non-goal oriented navigation. 
In contrast, our data-generation tool (\newdataset), aims to serve as a benchmark to goal-oriented navigation in the presence of humans. 
Since the data is fully collected in simulation using synthetic humans, our dataset and method avoid privacy concerns that might arise from using real human subjects.
Another benchmark on navigation \cite{unstuck-dinesh}, similarly to us, uses simulation for training, but is unsuitable for RGB-based visual navigation because humans in the scene have no visual texture and features, which are known to be important for closing the sim-to-real gap reliably~\cite{varol17_surreal}.
\SBnote{Game engines such as Unity and Unreal can also be used to create photorealistic data for learning navigation policies.
A key advantage of \newdataset is that it can be used with already existing, photorealistic static environment datasets, such as Habitat \cite{savva2019habitat} and Gibson \cite{xia2018gibson}, to generate training data with a minimal setup.}
\vspace{-0.5em}
\section{Problem Setup} \label{sec:formulation}
\vspace{-0.5em}
We study the problem of autonomous robot navigation in an \textit{a priori} unknown indoor space shared with a human whose trajectory is unknown to the robot. While robot state estimation and the effect of the robot trajectory on human motion are important problems, we assume in this work that the robot state can be estimated reliably and that the human expects the robot to avoid her. We model the vehicle as a three-dimensional, nonlinear system with dynamics (discretized for planning):
\begin{equation} \label{eqn:NumSimpleDyn}
\dot{x}^V = v^V\cos{\phi^V},\quad \dot{y}^V = v^V\sin{\phi^V},\quad \dot{\phi}^V = \omega^V\,,
\end{equation}
where $\state_{\time} := (x^V_{\time}, y^V_{\time}, \phi^V_{\time})$ is the state of vehicle, consisting of position $\pos^V_{\time} = (x^V_{\time}, y^V_{\time})$ and heading $\phi^V_{\time}$. The inputs (control) to the vehicle are $\ctrl_{\time} := (v^V_{\time}, \omega^V_{\time})$, consisting of speed $v^V_{\time}$ and turn rate $\omega^V_{\time}$ that are bounded within $[0, \bar{v}^V]$ and $[-\bar{\omega}^V, \bar{\omega}^V]$ respectively. 
The robot observes the environment through a forward-facing, monocular RGB camera mounted at a fixed height and oriented at a fixed pitch.
The goal of this paper is to learn control policies based on these images to go to a target position, $\pos^*_V = (x^*, y^*)$, specified in a global coordinate frame, while avoiding collision with the human, as well as any fixed obstacles.
\section{Model-Based Learning for Navigation Around Humans} \label{sec:approach}

\subsection{Learning-Based WayPoint for Navigation around Dynamic Humans (\small LB-WayPtNav-DH)} \label{subsec:lb_wayptnav}
Our approach, Learning-Based WayPoint for Navigation around Dynamic Humans (LB-WayPtNav-DH), uses two modules for navigation around humans: perception, and planning and control (see Appendix \ref{appendix_sec:arch:lb_wayptnav} for more details). 

\textbf{Perception Module:} \label{subsubsec:lb_wayptnav_perception_module}
The goal of the perception module is to analyze the image and provide a high-level plan for the planning and control module.
We implement the perception module using a Convolutional Neural Network (CNN) which inputs a 224 $\times$ 224 RGB image obtained from the onboard camera, the desired goal position in the robot's current coordinate frame, and the robot's current linear and angular speed.
The CNN outputs the robot's next desired state, or waypoint, $\hat{w_t}= (\hat{x_t}, \hat{y_t}, \hat{\theta_t})$.
The system is trained using an automatically generated expert policy (Sec. \ref{subsection:training_details}).

\textbf{Planning and Control Module:}
\label{subsubsec:lb_wayptnav_control_module}
Given the desired waypoint $\hat{w_t}$, the planning and control module generates a low-level plan and associated control inputs for the robot.
Since we run all computations onboard, we use computationally efficient spline-based trajectories to plan a smooth, efficient, and dynamically feasible (with respect to the dynamics in \eqref{eqn:NumSimpleDyn}) trajectory from the robot's current state to $\hat{w_t}$.
To track the trajectory, we design an LQR controller for the linearized dynamics around the trajectory. 
The controller is executed on the robot for a control horizon of $H$ seconds, at which point the robot receives a new image of the environment and repeats the process.

\subsection{Data Generation Procedure}\label{subsection:training_details}
We train the perception module entirely in simulation with self-supervision, using automatically generated RGB images and optimal waypoints as a source of supervision. 
The waypoints are generated so as to avoid the static obstacles and humans, and make progress towards the goal.
To generate these waypoints, we assume that the location of all obstacles is known during training time.
This is possible since we train the CNN in simulation; however, no such assumption is made during the test time. 
Under this assumption, we propose an MPC-based expert policy to generate realistic trajectories for humans, and subsequently, optimal robot trajectories and waypoints.
To obtain photorealistic images of the scene we develop the HumANav data-generation tool (Sec. \ref{subsection:dataset}). 

\textbf{MPC-Based Expert Policy.}\label{subsection:mpc_expert_policy}
To generate realistic human trajectories, we model the human as a goal-driven agent with state $\stateH$ and dynamics given by \eqref{eqn:NumSimpleDyn}. We additionally make the simplifying assumption that the human follows a piecewise constant velocity trajectory. 
This assumption is often used in human-robot interaction literature to obtain a reasonable approximation of human trajectories \cite{rudenko2019human}.

For the purposes of generating trajectory data for training the perception module, the human and robot are both modeled as cylindrical agents. To generate the training data, we first sample the start positions ($p_V^0$, $p_H^0$) and the goal positions ($p_V^*$, $p_H^*$) for the robot and the human respectively, as well as a unique identity for the human (body shape, texture, gender).
We then use receding horizon MPC to plan paths for both the robot and human for a horizon of $H_p$.
In particular, at time $t$, the human solves for the optimal trajectory $\textbf{z}_H^*$ that minimizes the following cost function
\begin{equation}
    J^{H}(\textbf{z}_H, \textbf{u}_H) = \sum_{i=t}^{t+H_p} J^{H}_i(\stateH_i, \ctrlH_i)
\end{equation}
\begin{multline}
    J^{H}_i(\stateH_i, \ctrlH_i) = \lambda_1^H d^{goal}_H(\stateH_i)^2 + \\ \lambda_2^H (\max\{0, d^{obs}_{cutoff}-d^{obs}(\stateH_i)\})^3
    + \lambda_3^H \norm{\ctrlH_i}
\end{multline}
Here $d^{goal}_H(\stateH_i)$ represents the minimum collision-free distance between $\stateH_i$ and the human goal position $p^*_H$ (also called Fast Marching Method (FMM) distance). 
$d^{obs}$ represents the signed distance to the nearest static obstacle. 
\SBnote{Using the signed distance (as opposed to the unsigned distance) ensures that the planning algorithm strictly prefers a trajectory that goes close to an obstacle (but not through it) compared to a trajectory that goes through the obstacle.} 
The obstacle cost is penalized only when the human is closer than $d^{obs}_{cutoff}$ to the nearest obstacle. 
The coefficients $\lambda_1^H, \lambda_2^H, \lambda_3^H$ are chosen to weight the different costs with respect to each other.

Given the optimal human trajectory for time horizon $[t, t+H_p]$, $\textbf{z}_H^*$, the robot optimizes for the waypoint, $\hat{w}_t$, such that the corresponding trajectory to that waypoint minimizes the following cost function:
\begin{equation}
    \label{eqn:vehicle_J}
    J^V(\textbf{z}_V, \textbf{u}_V) = \sum_{i=t}^{t+H_p} J^{V}_i(\state_i, \ctrl_i)
\end{equation}
\begin{multline} \label{eqn:vehicle_J_2}
    J^{V}_i(\state_i, \ctrl_i) = \lambda_1^V d^{goal}_V(\state_i)^2 + \lambda_2^V (\max\{0, d^{obs}_{cutoff}-d^{obs}(\state_i)\})^3 + \\\lambda_3^V (\max\{0, d^{human}_{cutoff}-d^{human}_i(\state_i)\})^3
\end{multline}
Similar to the human's cost function, $d^{goal}_V$ represents the collision-free distance to robot's goal, $p^*_V$,  $d^{obs}$ represents the signed distance to the nearest obstacle, and  $d^{human}_i$ represents the signed distance to the human at time $i$. 
The robot's distance to the human is only penalized when the robot and human are closer than $d^{human}_{cutoff}$ to each other. 
The coefficients $\lambda_1^V, \lambda_2^V, \lambda_3^V$ are chosen to weight the different costs with respect to each other.

Both the robot and human plan paths in a receding horizon fashion, repeatedly planning (for a horizon of $H_p$) and executing trajectories (for a horizon of $H$ where $H\leq H_p$) until the robot reaches its goal position. 
We then render the image seen at each of the robot’s intermediate states and save the corresponding pair $[(I_t, p^*_t, u^V_t), \hat{w}_t]$ for training using supervised learning.

\begin{remark}
\SBnote{Currently, we choose $\lambda_3^V$ to be high in \eqref{eqn:vehicle_J_2} -- this leads to navigation policies that are cautious around humans.
In future work, it will be interesting to vary the weights in \eqref{eqn:vehicle_J_2} to learn a suite of navigation policies ranging from cautious to aggressive.}
\end{remark}

\textbf{Data Sampling Heuristics:}\label{subsubsection:data_sampling_procedure}
We found that training on data with rich interaction between the robot and both static obstacles and humans was crucial to success in test scenarios, especially on our hardware setup; this includes episodes where the robot must navigate around chairs, through doorways, behind a slowly moving human, cut across a human's path, etc. To this end, we designed several heuristics to stimulate such interaction. \textit{First,} we choose the human's initial state, $p_H^0$, such that it lies approximately along the robot's optimal path to its goal position in the absence of the human. \textit{Second,} we penalize for proximity to the human \textit{only} when the human is visible in the robot's current RGB image. This facilitates downstream learning as it ensures the human is visible when information about the human is used for planning (see Sec. \ref{section:appendix:data_sampling} in supplementary Appendix for quantitative results on the importance of these sampling heuristics).

\subsection{The Human Active Navigation Data-Generation Tool (\newdataset)}\label{subsection:dataset}
\SBnote{The data generation procedure described in Sec. \ref{subsection:training_details} requires an environment whose map is known \emph{a priori} and  capabilities for creating dynamic environments with a desired human pose and identity.
Moreover, since future scenes themselves depend on the robot motion policy, we should be able to render realistic visuals of the environment, the humans, and their motion from \textit{any} robot viewpoint.
To the best of our knowledge, no existing simulation-based tool supports all these functionalities, so we created the Human Active Navigation Data-Generation Tool (\newdataset).}

\newdataset shown in Figure \ref{fig:front_figure}, is an active data-generation tool incorporating 6000 human meshes from the SURREAL dataset~\cite{varol17_surreal} and 7 indoor office building meshes from the SD3DIS dataset~\cite{armeni_cvpr16}.

The key component of \newdataset is a rendering engine that automatically fuses these meshes in order to allow a user to load a human, specified by gender, texture (clothing, skin color, facial features), and body shape, into a desired building, at a specified position, orientation, speed, and angular speed. 
Additionally, the user can manipulate the human pose and render photo-realistic visuals (RGB, disparity, surface normals, etc.) of the human and building from arbitrary viewpoints using a standard perspective projection camera.
Crucially, \newdataset renders images with visual cues relevant for path planning (\SBnote{e.g., the legs of a stationary or a very slow moving human will be closer together compared to those of a very fast moving human}), ensuring that visual cues for downstream planning are present in imagery.
Note that even though we use the SD3DIS dataset in \newdataset, our rendering engine is independent of the meshes used and textured meshes from any office buildings can be used.

Once we generate the human and robot trajectories as described in Sec. \ref{subsection:training_details}, we use \newdataset to render the RGB images along those trajectories. 
The rendered RGB images along with the optimal waypoints are used to train the CNN in our perception module with supervised learning.

\section{Simulation Experiments} \label{sec:sims}
We now present simulation and real-world experimental results to investigate the following two key questions:
(1) Can \metName effectively plan goal-driven trajectories in novel environments while reasoning about the dynamic nature of humans? (2) What are the merits of combining model-based control with learning for perceptual understanding of humans, compared to fully learning-based methods and purely geometry-based, learning-free methods?

\textbf{Simulation Setup:} Our simulation experiments are conducted using the \newdataset data-generation tool described in Section \ref{subsection:dataset}. Scans from 3 buildings and 4800 human identities are used to generate training data. 150 test episodes (start, goal position pairs) in a 4th \textit{held out} building and \textit{held out} human identities (texture, body shape, etc.) are used to evaluate all methods (see Appendix Section \ref{section:training_and_test_areas} for representative images of our training and test environments).
Train and test scenarios are sampled to stimulate rich interaction between the human and the robot as described in Section \ref{subsection:training_details}.

\textbf{Implementation Details:} We use a pretrained ResNet-50 to initialize the CNN-based perception module and finetune it using a MSE loss and ADAM optimizer with learning rate $10^{-4}$ and $L2$ weight decay coefficient of $10^{-6}$ on 125k data samples from \newdataset (more details in Appendix Sec. \ref{appendix_sec:arch}).
\begin{table*}
\centering
\resizebox{0.9 \linewidth}{!}{
\begin{tabular}{lccccc}
\textbf{Agent} & \textbf{Input} & \textbf{Success (\%)} & \textbf{Time Taken (s)} & \textbf{Acc ($m/s^{2}$)} & \textbf{Jerk ($m/s^{3}$)}\\ \midrule
Expert & Full map & 100 \\
\midrule
\vspace{.25em}
\textbf{Learning Based Methods} & & \\
\metName{} (ours) & RGB & \textbf{80.52} & \textbf{12.37 \textpm 2.02} & \textbf{.09 \textpm .02} & \textbf{.60 \textpm .15}  \\
\metNameCorl{} & RGB & 67.53 & 14.19 \textpm 2.79 & .10 \textpm .02 & .71 \textpm .13 \\
E2E & RGB & 52.60 & 14.46 \textpm 3.26 & .14 \textpm .02 & 3.71 \textpm .87\\
\midrule
\vspace{.25em}
\textbf{Mapping Based Methods} (memoryless) & & \\
Mapping-SH  & Depth & 76.63 & \textbf{12.02 \textpm 1.64} & .11 \textpm .03 & .75 \textpm .25\\
Mapping-WC & Depth + Human State & \textbf{94.81} & 12.08 \textpm 2.14 & .10 \textpm .03 & \textbf{.71 \textpm .21}\\
Dynamic Window Approach (DWA)  & Depth & 63.63 &  17.96 \textpm 7.43 & \textbf{.06 \textpm .02} & 2.57 \textpm .77\\
\bottomrule
\end{tabular}}
\caption{Performance of \metName (ours) and the baselines in simulation. Best results shown in bold.}
\label{table:simulation_success_rate}
\vspace{-2.1em}
\end{table*}

\textbf{Baselines:} We compare \metName with five baselines. \textit{\metNameCorl{}}: the CNN is trained on the SD3DIS dataset with no humans.  
\textit{Mapping-SH (Static Human)}: the known robot's camera parameters are used to project its current depth image to an occupancy grid (treating the human as any other static obstacle), which is then used for model-based planning.
\SBnote{\textit{Dynamic Window Approach (DWA)} \cite{fox1997dynamic}: takes the current depth information and the goal coordinates as inputs and outputs the optimal linear and angular velocity commands to be applied on the robot.
The optimal velocity is selected to maximize the robots clearance, maximize the velocity, and obtain the heading closest to the goal.}
\textit{End-to-End (E2E) learning}: CNN trained on the same data as \metName{}, but instead of a waypoint directly regresses to control commands \SBnote{(i.e. linear and angular velocity)} corresponding to the optimal robot trajectory.
\textit{Mapping-WC (Worst Case Human)}: same as Mapping-SH, but if the human is visible in the current frame, Mapping-WC plans a path around all possible future human behaviors assuming that the human's current state, $[x_t^H, y_t^H, \phi_t^H]$, is perfectly known and that the human moves at any speed in $[0, \bar{v}^H]$ for the entire planning horizon. 
We use a control horizon of $H=0.5s$ for fast replanning around humans.
\SBnote{Note that all of the presented methods are memoryless -- they do not have any visual memory and only use the current scene information for decision making.} 

\textbf{Metrics:} We compare the success rate across all methods. An episode is successful if the robot reaches within $0.5$ meters of its goal without colliding with a human or static obstacle. We further compare \metName and other methods on episode specific metrics computed over the subset of goals where \textit{all} methods succeed; we report the average time to reach the goal, average robot jerk, and acceleration (Acc) along the successful trajectories (lower is better). 

\begin{figure}
    \centering
    \includegraphics[trim={0cm, 1.25cm, 0cm, 0cm}, clip, width=.30\columnwidth]{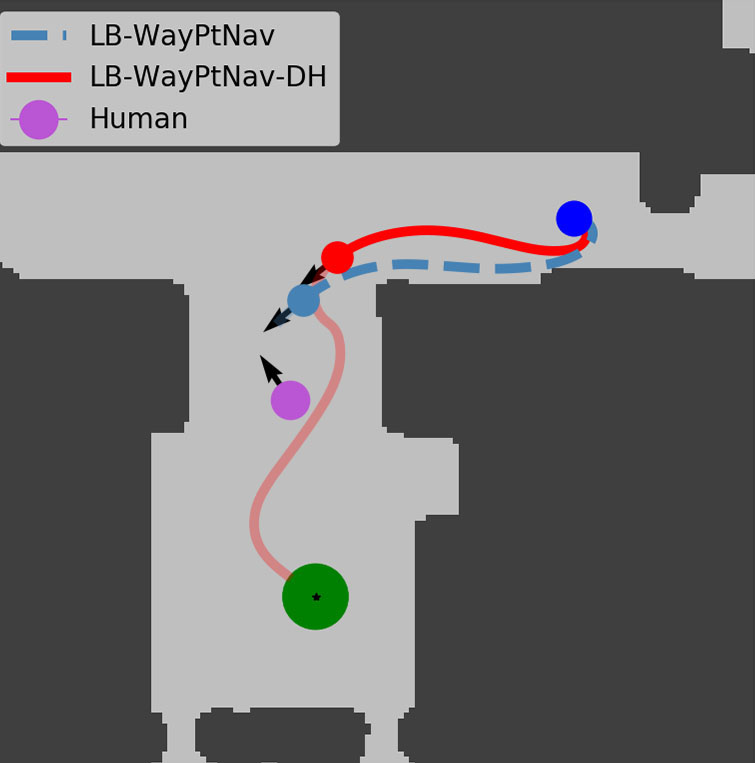}
    \includegraphics[width=.30\columnwidth]{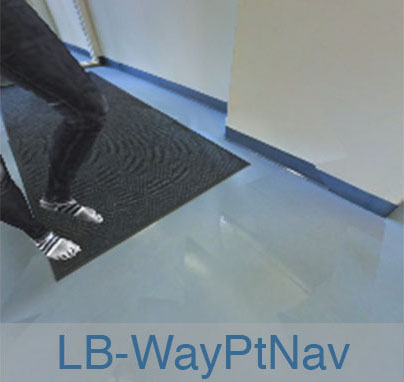}
    \includegraphics[width=.30\columnwidth]{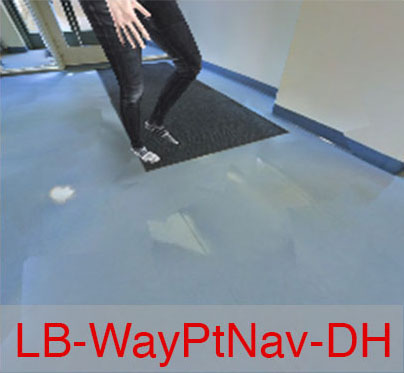}
    \caption{(left) The robot starts at the dark blue circle. Its goal it to move to the green goal region without colliding with static obstacles (dark gray) or humans (magenta). \metNameCorl follows the light-blue, dashed trajectory until the light blue dot, planning a path to the right of the human (in its direction of motion), leading to collision.  \metName follows the red trajectory until the red circle, planning a trajectory (transparent red) to the left the of the human which accounts for the its future motion, and ultimately leads to success. (middle \& right) Corresponding RGB images seen by the robot.}
    \vspace{-1.8em}
    \label{fig:top_view_comparison}
\end{figure}
\subsection{Results}

\textbf{Comparison with \metNameCorl{}:} \metName reaches the goal on average $13\%$ more than \metNameCorl{} (Table \ref{table:simulation_success_rate}). As expected, \metNameCorl tends to fail in scenarios where anticipating future human motion plays a pivotal role in planning a collision-free path. 
\metNameCorl takes a greedy approach in such scenarios, treating the human like any other static obstacle, ultimately leading to a collision with the human.
In Fig. \ref{fig:top_view_comparison} we analyze one such test scenario.  

\textbf{Comparison with End-to-End learning:}
Our findings (Table \ref{table:simulation_success_rate}) are consistent with results observed in literature for static environments \cite{bansal2019combining, kaufmann2018deep} -- the use of model-based control in the navigation pipeline significantly improves the success rate of the agent as well as the overall trajectory efficiency and smoothness (see the \textit{Jerk} column in Table \ref{table:simulation_success_rate}).
We note that E2E learning particularly fails in the scenarios where a precise control of the system is required, such as in narrow hallways or openings, since even a small error in control command prediction can lead to a collision in such scenarios.

\textbf{Comparison with Mapping-SH:} 
\SBnote{Mapping-SH has access to the privileged information -- the ground-truth depth (and consequently occupancy) of the scene; hence, it can avoid static obstacles perfectly.
Despite this, Mapping-SH fails in about 
$23\%$ of navigation scenarios.
This is because it treats the human as a static obstacle and reacts to them, failing in the scenarios where a reactive planner that does not take into account the dynamic nature of the human is not sufficient to avoid a collision. 
In contrast, \metName succeeds on $58.33\%$ of these scenarios, indicating that it can reason about the dynamic nature of the human.}

\SBnote{Despite a good success rate of LB-WayPtNav-DH on the above scenarios, it only slightly outperforms Mapping-SH overall. 
That is because \metName \textit{learns} to avoid collision with both static obstacles and dynamic humans based on a RGB image, and as a result, its failure modes include collision with static obstacles as well.
In contrast, Mapping-SH has access to the perfect geometry of the scene and can avoid static obstacles perfectly.
Mapping-SH is also approximately $9\%$ faster than \metName on the goals where both methods succeed because it can leverage the scene geometry to plan optimal paths that barely avoid the human.
\metName, on the other hand, is trained to take conservative trajectories which avoid the human's potential future behavior.} However, since real-world depth sensors are neither perfect nor have an unlimited range, we see a noticeable drop in the performance of Mapping-SH in real-world experiments as discussed in Sec. \ref{sec:exp}.
In contrast, \metName is trained to be robust to sensor noise and exhibits similar error profiles on real and synthetic imagery.

\begin{figure}
    \centering
    \includegraphics[width=.3 \columnwidth]{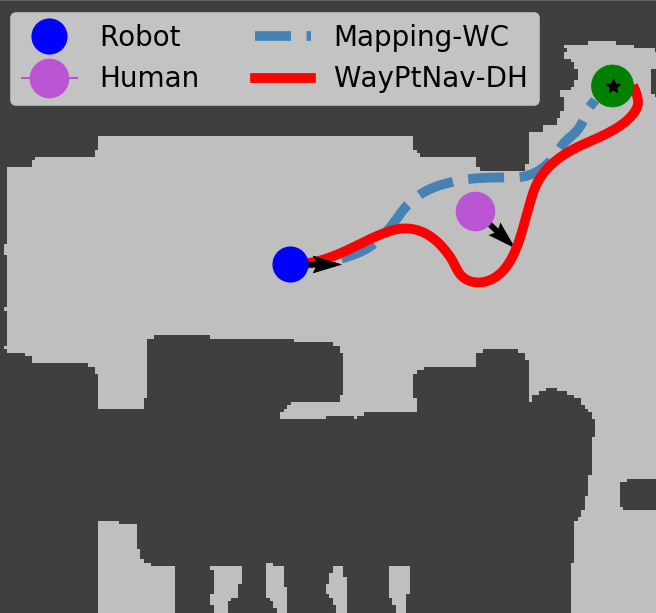}
    \includegraphics[width=.3 \columnwidth]{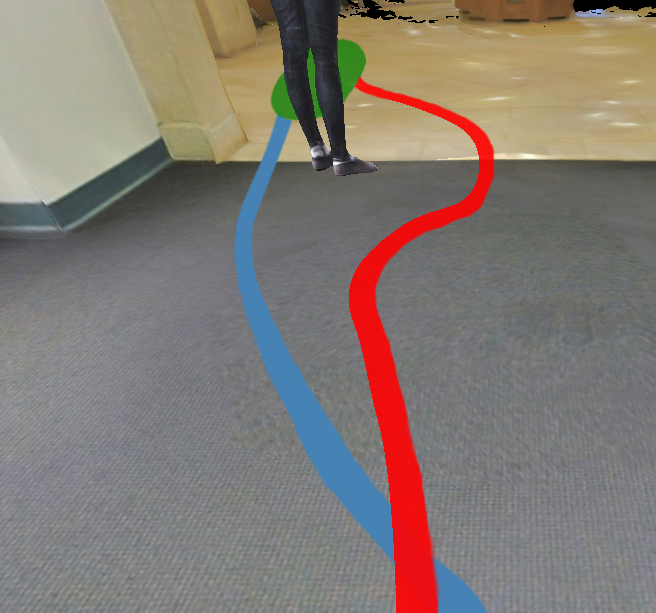}
    \caption{Top view of the trajectories taken by Mapping-WC and \metName from the same state and the corresponding RGB image with the trajectories superimposed. Mapping-WC reaches the goal faster than \metName as it has access to precise geometry of the scene and the human state and thus plans a path between the human and the wall which narrowly avoids collision. \metName, on the other hand, takes a more cautious path as it does not have access to the human state.}
    \vspace{-2.0em}
    \label{fig:mapping_wc_efficiency}
\end{figure}

\textbf{Comparison with Mapping-WC:}
Mapping-WC unsurprisingly achieves near perfect ($95\%$) success as it assumes perfect depth and human state estimation. Mapping-WC fails ($5\%$) due to the receding horizon nature of its MPC planner, which might lead the robot to a future state from which it cannot avoid collision.

Interestingly, we found that in many cases, Mapping-WC reaches the goal faster than \metName (Table \ref{table:simulation_success_rate}) by exploiting precise geometry of the scene and human, taking an aggressive trajectory which barely avoids collision with the human (see Fig. \ref{fig:mapping_wc_efficiency} for example).
\SBnote{We next compare \metName and Mapping-WC on the scenarios where Mapping-SH fails to successfully reach the goal. 
On these scenarios, \metName reaches the goal on average $6\%$ faster than Mapping-WC. 
This is not surprising as the failure of Mapping-SH indicates that it is important to account for the dynamic nature of the human to successfully reach the goal in these scenarios. 
However, as expected, Mapping-WC takes overly conservative paths in these scenarios, planning a path that avoids \emph{all} possible human trajectories regardless of their likelihood. 
In contrast, \metName is trained to reason about the human's likely trajectory and thus plans more efficient paths.}

Mapping-WC performance is also affected by noise in human state estimation.
To quantify this, we add zero-centered, uniform random noise to $[x_t^H, y_t^H, \phi_t^H]$ in Mapping-WC.
As a result, the success rate of Mapping-WC drops by $7\%$, indicating the challenges associated with this approach, especially when human state needs to be inferred from a monocular RGB image.

\textbf{Comparison with DWA:} 
\SBnote{Similar to Mapping-SH, DWA treats the human as a static obstacle and takes a greedy strategy to avoid them. This often leads to situations where the robot tries to avoid the human by moving in its direction of motion, ultimately resulting in a collision with the human.
Interestingly, despite being qualitatively similar, Mapping-SH significantly outperforms DWA. 
This is because DWA plans piecewise constant linear and angular speed profiles in order to make sure that the planning can be performed efficiently. However, this comes at a tradeoff in the agility of the robot, causing it to get stuck in tight corners and narrow openings. Mapping-SH on the other hand uses MPC to plan dynamically feasible, spline-based trajectories that are more agile and lead to continually varying speed profiles.
}

\begin{figure}
    \centering
    \includegraphics[trim={0cm, .8cm, 0cm, 0cm}, clip, width=.30\columnwidth]{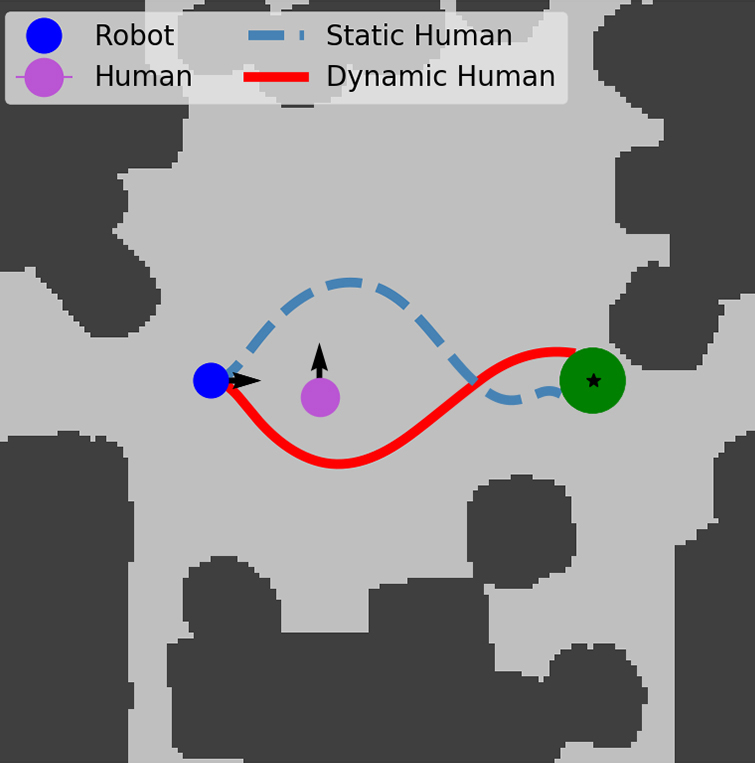}
    \includegraphics[width=.30\columnwidth]{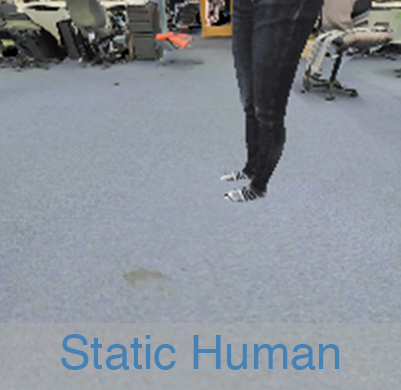}
    \includegraphics[width=.323 \columnwidth]{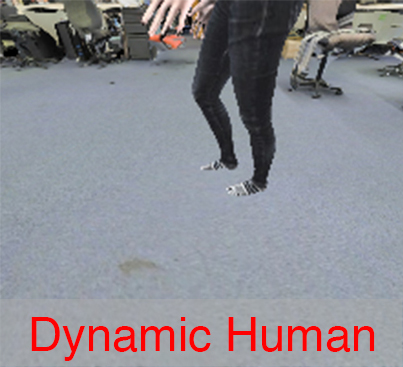}
    \caption{Top view of the trajectories taken by \metName from the same state with a static human (light blue, dashed line) and a dynamic human (red, solid line), and the corresponding RGB images. 
    \newdataset enables \metName to leverage cues, such as spread of humans legs and direction of human toes, to infer that the left RGB image likely corresponds to a static human and the right one to a moving human.}
    \vspace{-2.0em}
    \label{fig:legs_apart_visual_cues}
\end{figure}

\textbf{Learned Navigational Cues and Importance of Photorealism:}
We designed \newdataset such that relevant visual cues for navigation are rendered in imagery; i.e. a human's legs will be spread apart if they are moving quickly and will stay closed if they are not moving. 
\metName is able to incorporate these visual cues to anticipate future human motion and accordingly plan the robot's trajectory (Fig. \ref{fig:legs_apart_visual_cues}).

\begin{figure}[ht]
    \centering
    \includegraphics[width=.3 \columnwidth]{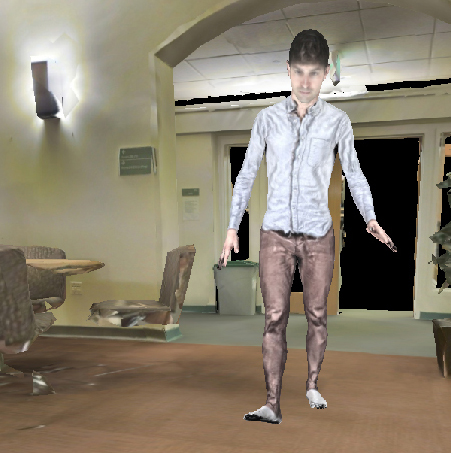}
    \includegraphics[width=.3 \columnwidth]{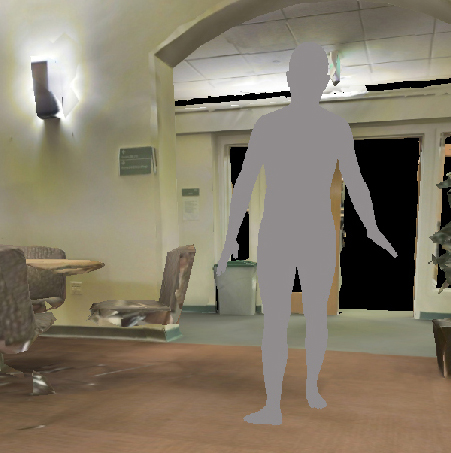}
    \caption{\metName trained on images from \newdataset with realistic textures (clothing, hair, skin color, facial features) (left) leads to a better generalization than training on human figures with gray textures (right).}
    \label{fig:gray_humans}
\end{figure}

To understand the importance of photorealistic images, we also trained \metName on images of humans that are colored in gray (see Fig. \ref{fig:gray_humans}). 
Consequently, we see a drop of $6\%$ in the success rate, indicating that training \metName with photorealistic textures (clothing, skin color, hair color, etc.) generalizes better to novel humans.

\begin{figure}
    \centering
    \begin{subfigure}{1.0 \columnwidth}
    \centering
    \includegraphics[width=.3 \columnwidth]{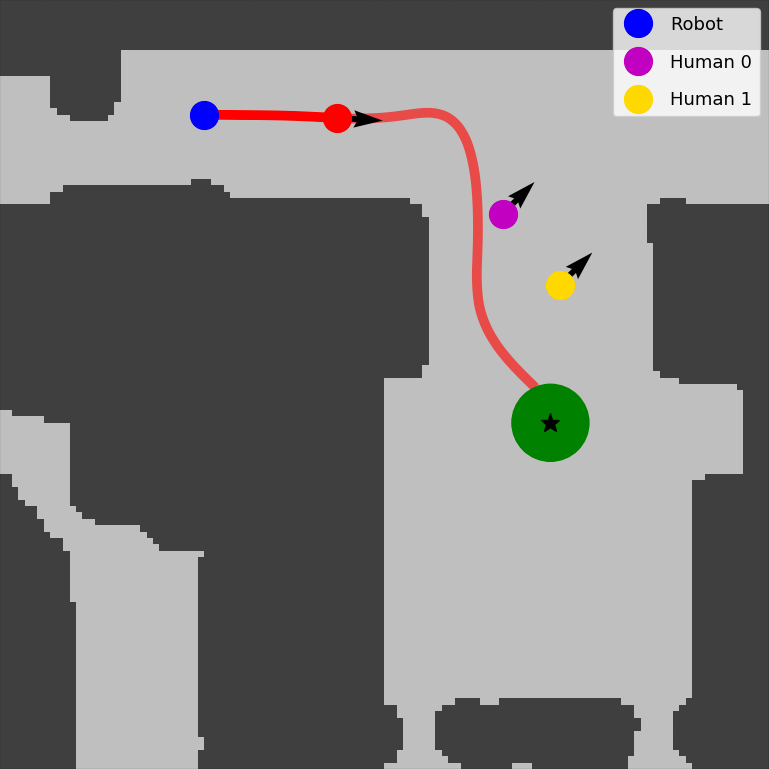}
    \includegraphics[width=.3 \columnwidth]{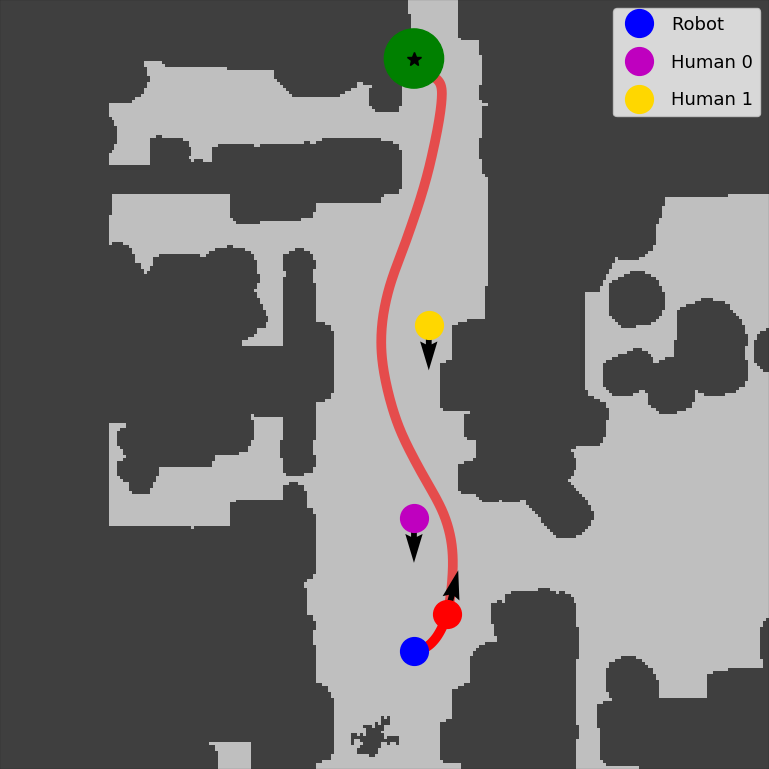}
    \includegraphics[width=.3 \columnwidth]{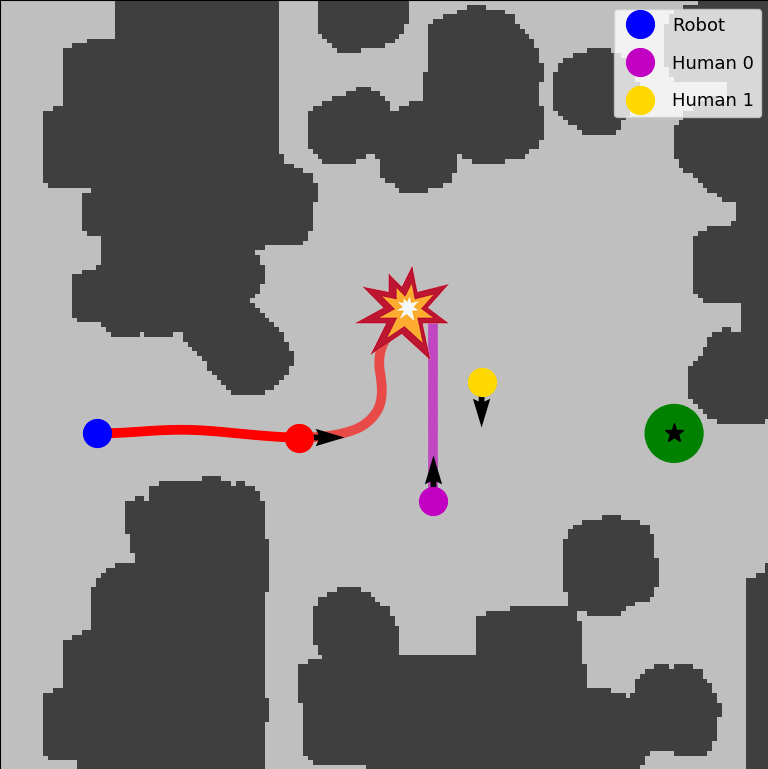}
    \end{subfigure}
    \par\smallskip
    \begin{subfigure}{1.0 \columnwidth}
    \centering
    \includegraphics[width=.3 \columnwidth]{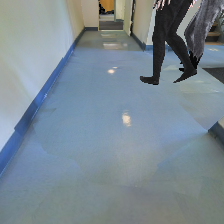}
    \includegraphics[width=.3 \columnwidth]{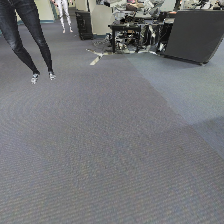}
    \includegraphics[width=.3 \columnwidth]{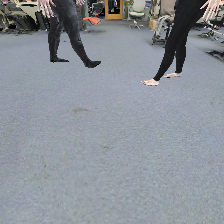}
    \end{subfigure}
    \caption{Navigation around multiple humans. \metName successfully turns a corner while avoiding two humans walking side by side (left) and navigates a long hallway with multiple humans walking down the hallway (middle). (right) \metName attempts to traverse a room, crossing the path of two different humans that are moving in opposing directions and is unable to reason about the future trajectory of both humans simultaneously which ultimately leads to a collision.}
    \vspace{-2.0em}
    \label{fig:multiple_humans}
\end{figure}

\textbf{Navigation Around Multiple Humans:}
\metName is trained on environments with a single human; however, we find that it can generalize to settings with multiple humans (Fig. \ref{fig:multiple_humans}).
\metName is able to successfully navigate around multiple humans walking side by side or separately in a narrow hallway. 
We hypothesize that \metName succeeds in these scenarios as it reduces the multi-human avoidance problem to a single human avoidance problem (i.e. by treating both humans as a single large "meta-human" in the first scenario and by solving two smaller, single-human avoidance problems in the second scenario). The third scenario, on the other hand, is specifically designed to test whether \metName can reason about multiple, distinct future human trajectories at once. \metName, struggles to accurately infer both humans' future motion, and thus collides. In fact, the same scenario, when run without the second human, leads to \metName successfully reaching the goal.

\textbf{Failure Modes:} \metName successfully navigates around dynamic and static obstacles in novel environments, however it is primarily limited in its ability to recognize and predict the \textit{long-term} motion of humans. These issues are tightly coupled with the robot's reactive nature (uses \emph{only} the current RGB image) and limited field of view (forward facing camera) as humans may approach the robot from outside or on the fringe of its field of view.
\section{Hardware Experiments} \label{sec:exp}
We directly deploy the LB-WayPtNav-DH framework, trained in simulation, onto a Turtlebot 2 hardware platform without any finetuning or additional training.
Our algorithm is tested in four experimental setups across two never-before-seen buildings (see Fig. \ref{fig:experiment_scenarios} in Appendix for some representative images). 
Importantly, we note that our robot has only been trained on synthetic humans from the SURREAL dataset \cite{varol17_surreal}, constrained to piecewise constant velocity trajectories. 
Humans in our experiments, however, do not have such dynamical constraints. 
For robot state measurement, we use the Turtlebot’s encoder based odometry. 

\begin{figure*}
    \vspace{-1.0 em}
    \centering
    \includegraphics[width=0.96\textwidth]{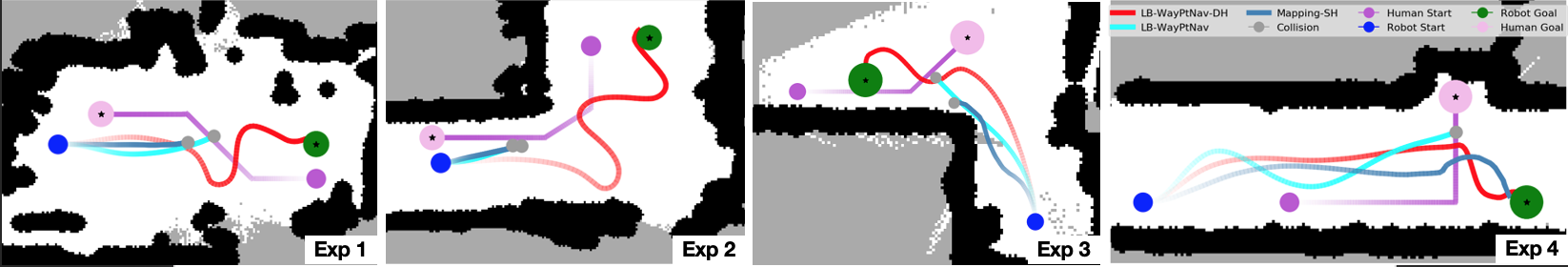}
    \vspace{-0.5em}
    \caption{\SBnote{Robot trajectories corresponding to different methods in the hardware experiments: red (\metName), cyan (\metNameCorl), and blue (Mapping-SH). The human trajectory is shown in purple. Earlier timesteps are shown more transparent. 
     \metName is able to anticipate and react to the human motion to avoid a collision, even if it means diverging from the optimal path to the goal.}}
    \vspace{-2.0em}
    \label{fig:experiment_settings}
\end{figure*}

\SBnote{The experiments are designed to evaluate whether the robot has learned to reason about the dynamic nature of humans (see Fig. \ref{fig:experiment_settings}). 
In Experiment 1, the human walks parallel to the robot but in the opposite direction; however, the human suddenly takes a turn towards the robot, requiring it to anticipate the human behavior to avoid a collision. 
In Experiment 2, the robot and the human move in opposite directions, but cross each other near a tight corner, requiring the robot to take a cautious trajectory around the human to avoid a collision.
In Experiment 3, the two agents are walking in perpendicular directions. For a successful collision avoidance, the robot needs to reason about the direction of human motion and react to it, while also avoiding a collision with the corner wall.
In Experiment 4,  the robot is moving at its full speed behind a human in a hallway. However, the human suddenly stops and starts moving perpendicular to the hallway.} 

For each experimental setting, we conduct five trials each for \metName, \metNameCorl, and Mapping-SH (a total of 20 experiments per method).
We do not compare to End-To-End or Mapping-WC on our hardware setup as the simulation performance of End-To-End is already very low and Mapping-WC requires access to the ground truth state information of the human, which was not reliable using our narrow field-of-view monocular RGB camera.
The experiment videos are available on the project website\footnote{Project website: \href{https://smlbansal.github.io/LB-WayPtNav-DH/}{\textcolor{blue}{https://smlbansal.github.io/LB-WayPtNav-DH/}}}.

\textbf{Comparison With \metNameCorl:} 
\metNameCorl succeeds in only 3 trials out of 20. 
In all experiments \metNameCorl attempts to avoid the human, treating it as a static obstacle; however, the human advances towards the robot before it can correct course, leading to a collision with the human. 
This is unsurprising as this method is trained purely on static obstacles and these experiments are specifically designed to test the agent's understanding of the dynamic nature of humans.

\textbf{Comparison With Mapping-SH:} To implement Mapping-SH on the Turtlebot, we project the robot's current depth image onto an occupancy grid on the ground plane using RTAB-MAP package. 
Similar to \metNameCorl, Mapping-SH avoids the human by treating them as static obstacles, leading to its poor performance in our hardware experiments (it succeeds in 7/20 trials).
Performance of Mapping-SH is further impacted by its over reliance on the geometry of the scene.
\SBnote{This is particularly evident in Experiment 3 where the robot tries to sneak through the narrow gap between the human and the wall, but ends up failing due to the inevitable noise in the real-world depth sensor.}

Given the reactive nature of Mapping-SH and lack of understanding of the dynamic nature of the human, when Mapping-SH does succeed it does so by executing a last-resort, aggressive turn or stop to avoid imminent collision with the human. 
\SBnote{This is evident in Experiment 4 where the robot first moves in the direction of motion of the human, but later corrects its course by stopping and taking a right turn towards the goal.}

\textbf{Performance of \metName:} \metName succeeds in all 20 trials by exhibiting behavior which takes into account the dynamic nature of the human. 
These results demonstrate the capabilities of a learning algorithm trained \emph{entirely} in simulation on the \newdataset dataset to generalize to navigational problems in real buildings with real people.

In Experiment 1 (Fig. \ref{fig:experiment_settings}) \metName navigates around the human by moving contrary to its direction of motion, which allows it to reliably avoid collision. 
\metNameCorl and Mapping-SH, however, treat the human as a static obstacle and attempt to avoid it by moving in its direction of motion. 
\SBnote{In Experiment 2, LB-WayPtNav-DH is able to learn that to avoid a collision with the human, it should not attempt to cross the human’s path and instead walk parallel to the human until it passes the human. In contrast, \metNameCorl and Mapping-SH exhibit greedy behavior and attempt to cut the human path in hope for a shorter path to the goal, ultimately leading to a collision.} \SBnote{In Experiment 3, LB-WayPtNav-DH avoids the human by turning in the opposite direction to the human motion; however, at the same time, it slows down to avoid a collision with the wall ahead. Once the human passes, the robot turns away from the wall to reach its goal position.} \SBnote{Finally, in Experiment 4, LB-WayPtNav-DH is successfully able to avoid a collision with the human by coming to a complete stop and waiting for the human to pass. Once the human passes, the robot navigates to its goal.}

\SBnote{Even though successful at avoiding humans and reaching the goal, we notice that \metName exhibits some oscillations in the robot trajectory, leading to sub-optimal trajectories to the goal.
These oscillations are primarily caused by the narrow FOV of the camera ($\approx 50$ degrees) in our hardware experiments. Since \metName relies only on a monocular RGB image, a narrow FOV limits its reasoning abilities about obstacles in the environment and the optimal path to the goal, especially because the robot is operating in an unknown environment. This leads to prediction of sub-optimal waypoints, and consequently, sub-optimal trajectories to the goal.}
\section{Conclusion and Future Work} \label{sec:conclusion}
We propose LB-WayPtNav-DH, a framework that combines a learning-based perception module and a model-based planning module for autonomous navigation in \textit{a priori} unknown indoor environments with humans.
To train the perception module in LB-WayPtNav-DH, we also create a photorealistic data-generation tool, \newdataset, that can render rich indoor environment scenes with humans. \newdataset consists of synthetic humans and can be interfaced with fully automatically, avoiding privacy and logistic difficulties present when working with real human subjects. 
We demonstrate that LB-WayPtNav-DH trained on \newdataset can successfully learn to navigate around humans and transfer the learned policies from simulation to reality.

In future work, it would be interesting to learn richer navigation behaviors in more complex and crowded scenes with multiple humans.
\SBnote{We use one of the simplest models of human prediction that exists, in order to train the network. There is a wealth of ongoing work on developing predictive models of humans and how they interact with autonomy, which could be considered in the proposed framework in the future.} 
\SBnote{Currently, we learn navigation policies using monocular RGB images. It would be interesting to extend LB-WayPtNav-DH to consider other visual modalities, such as depth images.} \SBnote{Finally, dealing with noise in robot state estimation and adding visual memory for optimal, long-range navigation will be another interesting future direction.}

\bibliographystyle{IEEEtran}
\bibliography{references}

\clearpage
\section{APPENDIX}
\begin{figure}
    \centering
    \includegraphics[width=1.0\columnwidth]{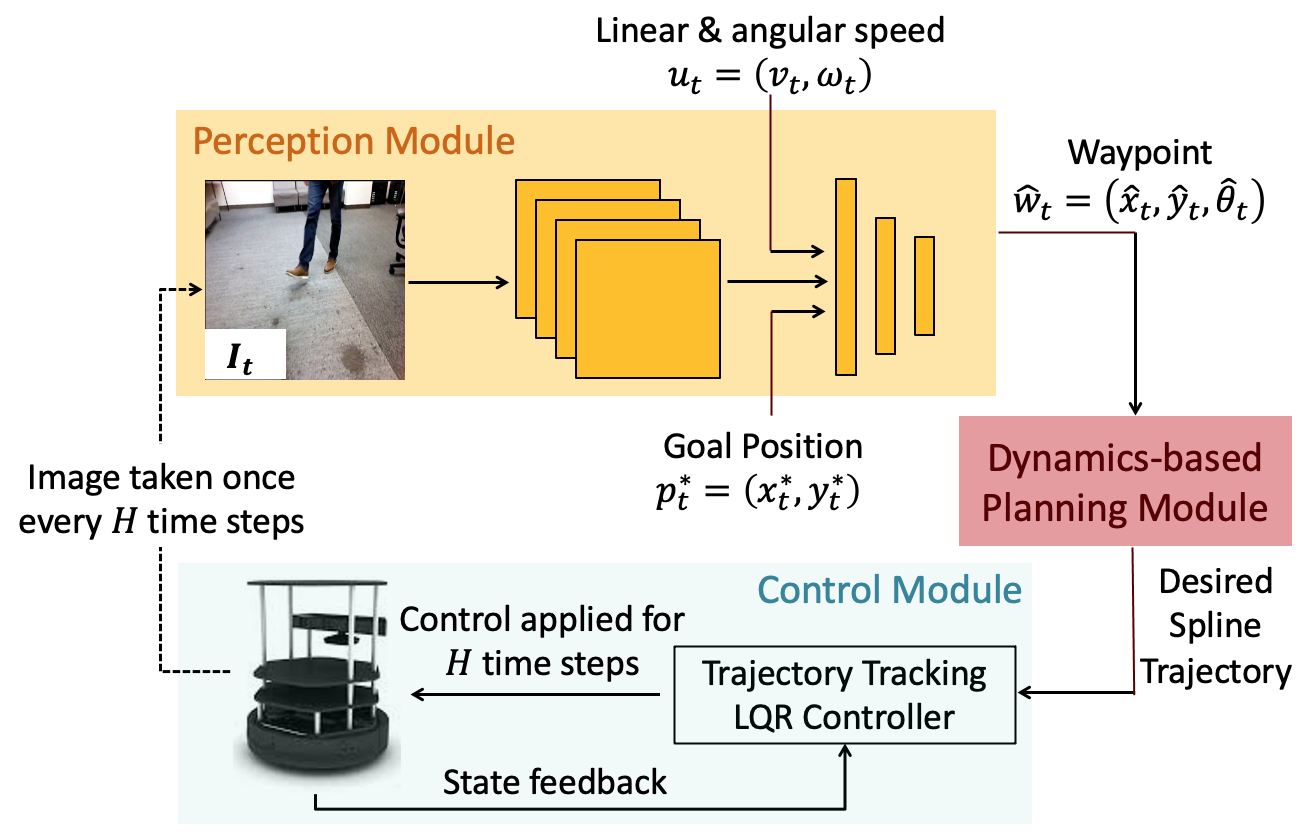}
    \caption{The LB-WayPtNav-DH framework to navigate in a priori unknown dynamic environments around humans.}
    \label{fig:lb_wayptnav}
\end{figure}

\SBnote{\subsection{Learning-Based WayPoint for Navigation around Dynamic Humans (\small LB-WayPtNav-DH)} \label{appendix_sec:arch:lb_wayptnav}
Our approach, Learning-Based WayPoint for Navigation around Dynamic Humans (LB-WayPtNav-DH), combines a learning-based perception module with a dynamics model-based planning and control module for navigation in novel dynamic environments (Fig. \ref{fig:lb_wayptnav}). 
We give a brief overview of the perception and planning/control modules of LB-WayPtNav-DH here.} 

\SBnote{\subsubsection{Perception Module} \label{subsubsec:lb_wayptnav_perception_module_appendix}
The goal of the perception module is to analyze the image and provide a high-level plan for the planning and control module.
The perception module leverages a Convolutional Neural Network (CNN), $\psi$ which inputs $I_t$, a 224 $\times$ 224 RGB image obtained from the onboard camera; $p_t^*$, the desired goal position in the robot's coordinate frame; and $u_t^V$, the robot's current linear and angular speed. 
The CNN is trained to output the robot's next desired state, or waypoint, $\hat{w_t}=(\hat{x_t}, \hat{y_t}, \hat{\theta_t})=\psi(I_t, p_t^*, u_t^V)$.
The CNN is trained using an automatically generated expert policy (Sec. \ref{subsection:training_details}).}

\SBnote{\subsubsection{Planning and Control Module}
\label{subsubsec:lb_wayptnav_control_module_appendix}
Given the desired waypoint $\hat{w_t}$, the planning and control module generates a low-level plan and associated control inputs for the robot.
Since we run all computations onboard, we use computationally efficient spline-based trajectories to plan a smooth, efficient, and dynamically feasible (with respect to the dynamics in \eqref{eqn:NumSimpleDyn}) trajectory from the robot's current state to $\hat{w_t}$.
To track the trajectory, we design an LQR controller for the linearized dynamics around the planned spline trajectory.
The controller is executed on the robot for a control horizon of $H$ seconds, at which point the robot receives a new image of the environment and repeats the process.}

\subsection{Network Architecture and Training Details} \label{appendix_sec:arch}
We train \metName{} and E2E learning on 125K data points generated by our expert policy (Section \ref{subsection:mpc_expert_policy}). 
All our models are trained using TensorFlow with a single GPU worker. 
We use MSE loss on the waypoint prediction (respectively on the control command prediction for E2E learning) for training the CNN in our perception module (respectively for E2E learning).
We use Adam optimizer to optimize the loss function with a batch size of 64.
We train both networks for 35K iterations with a constant learning rate of $10^{-4}$ and use a weight decay of $10^{-6}$ to regularize the network.
We use ResNet-50, pre-trained for ImageNet Classification, as the backbone for our CNN.
We remove the top layer, and use a downsampling convolution layer, followed by 5 fully connected layers with 128 neurons each to regress to the optimal waypoint (or control commands for E2E learning).
The image features obtained at the last convolution layer are concatenated with the egocentric target position and the current linear and angular speed before passing them to the fully connected layers.

During training, the ResNet layers are finetuned along with the fully connected layers to learn the features that are relevant to the navigation tasks.
We use standard techniques used to avoid overfitting including dropout following each fully connected layer except the last (with dropout probability $20\%$), and data augmentation such as randomly distorting brightness, contrast, adding blur, perspective distortion at training time.
Adding these distortions significantly improves the generalization capability of our framework to unseen environments.

\subsection{Importance of Data Sampling Heuristics} \label{section:appendix:data_sampling}
To understand the importance of our data sampling procedure, we train an additional baseline \textit{\metNameFov{}}.
In this baseline, the CNN is trained to predict waypoints which always avoid the human, regardless of whether the human is visible in the robot's current image or not.
To generate optimal waypoints for training the CNN, the robot cost function always penalizes the proximity with a human even when the human in not within the field-of-view (FOV) at the current time.
The results are presented in Table \ref{table:appendix:comparision_with_lb_wayptnav_fov}.

\begin{table}[ht]
\caption{Comparison between \metName (ours) and \metNameFov{} methods on 150 test episodes. 
Average time taken, jerk, and acceleration numbers are reported on the scenarios where both methods succeed.}
\label{table:appendix:comparision_with_lb_wayptnav_fov}
\centering
\resizebox{1.0\linewidth}{!}{
\begin{tabular}{lcccccc}
\toprule
\textbf{Agent} &  \textbf{Input} & \textbf{Success (\%)} & \textbf{Time Taken (s)} & \textbf{Acc ($m/s^{2}$)} & \textbf{Jerk ($m/s^{3}$)} \\ \midrule
\metName{} (ours) & RGB & 80.52 & 12.94 \textpm 3.06 & .09 \textpm .02 & .64 \textpm .13\\
\metNameFov{} & RGB & 68.18 & 13.57 \textpm 3.52 & .09 \textpm .02 & .66 \textpm .13\\
\bottomrule
\end{tabular}}
\end{table}

\metName reaches the goal on average $12\%$ more than \metNameFov{} and on average $5\%$ faster than \metNameFov{}. 
This indicates that restricting our expert to choose waypoints only considering information within its current field of view, as described in \ref{subsubsection:data_sampling_procedure}, facilitates downstream learning and ultimately the performance for \metName.
Intuitively, since the perception module is reactive, it has limited capabilities to reason about the human motion when the human is not in robot's FOV. 
Thus, reasoning about the human motion when the human is not within the FOV can overconstrain the learning problem.
In future, we will explore adding memory to the CNN (such as using LSTM or RNN) that can overcome some of these challenges.

\subsection{Training and Test Areas} \label{section:training_and_test_areas}
\begin{figure}[ht!]
\centering
\begin{subfigure}[b]{\columnwidth}
\centering
\begin{subfigure}[b]{0.24\columnwidth}
\centering
  \includegraphics[width=\columnwidth]{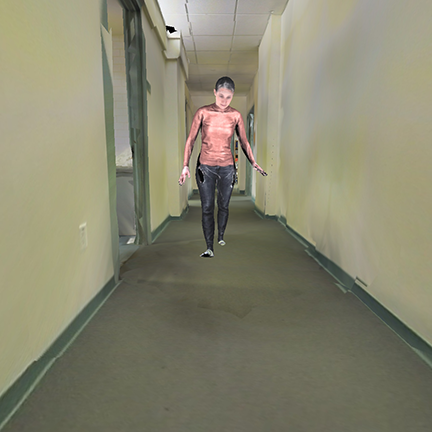}
\end{subfigure}%
\hfill
\begin{subfigure}[b]{0.24\columnwidth}
\centering

  \includegraphics[width=\columnwidth]{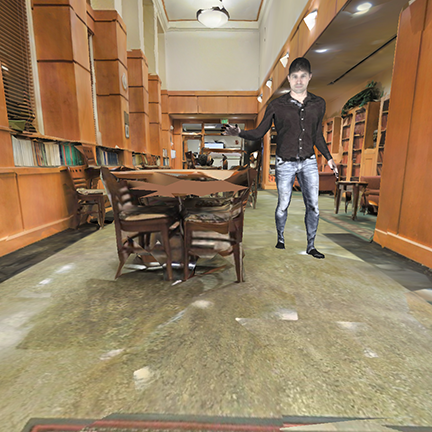}
\end{subfigure}
\hfill
\begin{subfigure}[b]{0.24\columnwidth}
\centering
  \includegraphics[width=\columnwidth]{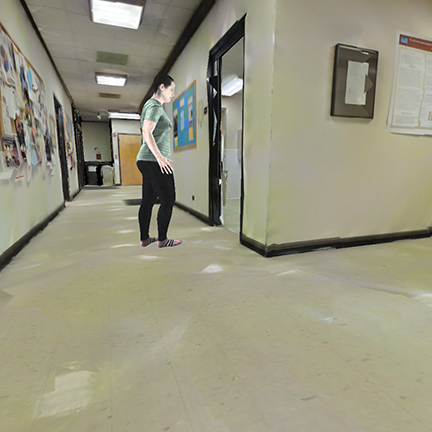}
\end{subfigure}
\hfill
\begin{subfigure}[b]{0.24\columnwidth}
\centering
  \includegraphics[width=\columnwidth]{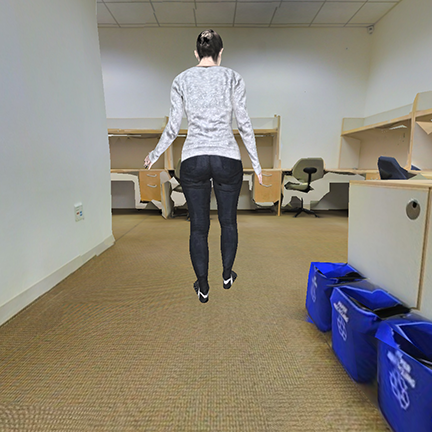}
\end{subfigure}
\caption{Sample training environments}
\end{subfigure}

\begin{subfigure}[b]{\columnwidth}
\centering
    \begin{subfigure}[b]{0.24\columnwidth}
    \centering
      \includegraphics[width=\columnwidth]{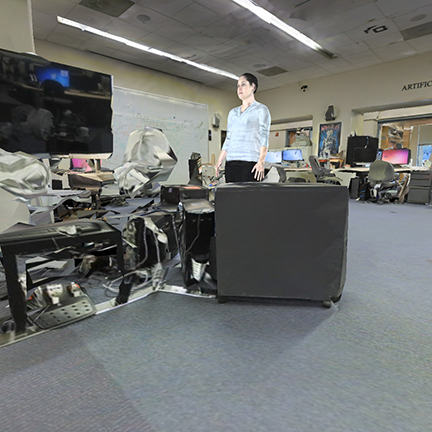}
    \end{subfigure}%
    \hfill
    \begin{subfigure}[b]{0.24\columnwidth}
    \centering
      \includegraphics[width=\columnwidth]{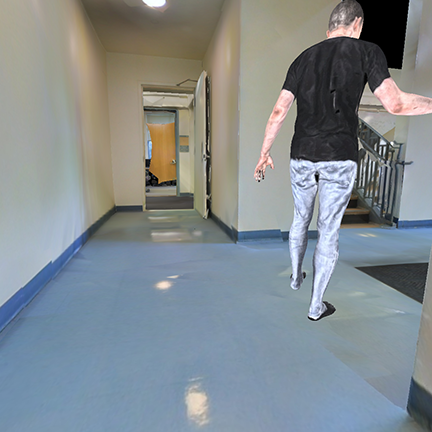}
    \end{subfigure}
    \hfill
    \begin{subfigure}[b]{0.24\columnwidth}
    \centering
      \includegraphics[width=\columnwidth]{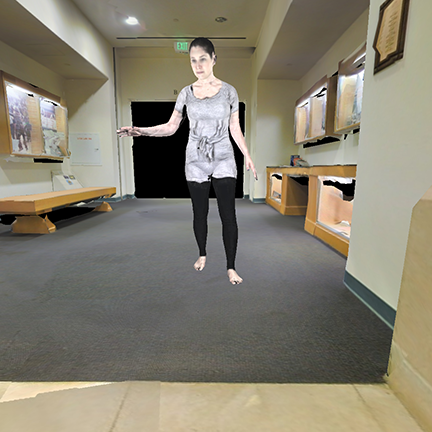}
    \end{subfigure}
    \hfill
    \begin{subfigure}[b]{0.24\columnwidth}
    \centering
      \includegraphics[width=\columnwidth]{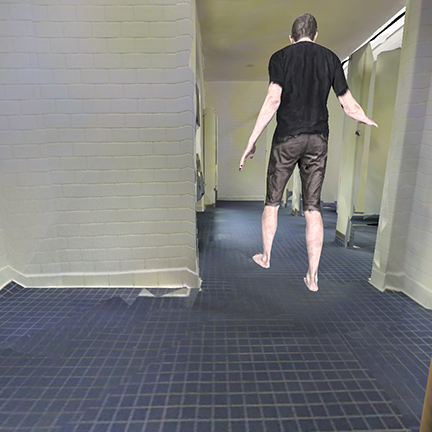}
    \end{subfigure}
    \caption{Sample test environments}
\end{subfigure}
\caption{Representative images from training and testing scenarios using the \newdataset data-generation tool. The buildings used at training and test time are visually dissimilar and have substantially different layouts. We also keep a held-out set of human identities for our test scenarios. \metName is able to generalize well to novel environments with never-before-seen humans at test time.}
\label{fig:training_test_areas}
\end{figure}

Training and testing in simulation is conducted using the \newdataset data-generation tool. Rendered RGB images from our training and testing environments are shown in Figure \ref{fig:training_test_areas}. 
Even though both the training and the test environments are indoor office spaces, their layout and appearance differ significantly, but \metName adapts well to the new environments. 

\begin{figure}[h!]
\centering
\begin{subfigure}[b]{0.48\columnwidth}
\centering
  \includegraphics[width=\columnwidth]{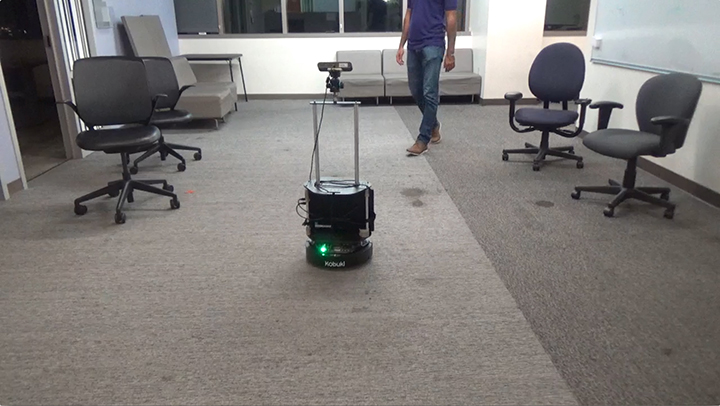}
\end{subfigure}%
\hfill
\begin{subfigure}[b]{0.48\columnwidth}
\centering

    \includegraphics[width=\columnwidth]{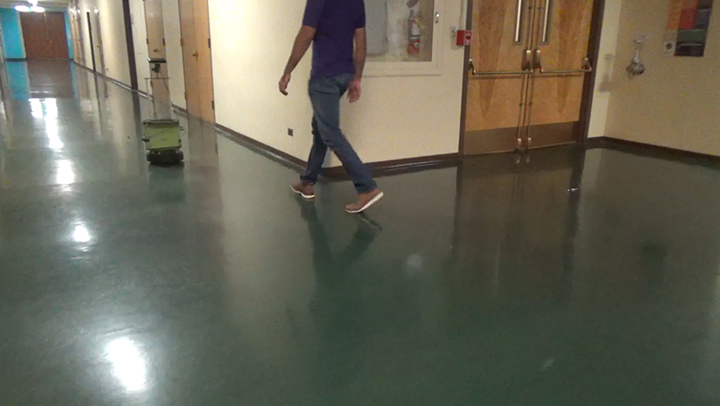}
\end{subfigure}%
\newline
\newline
\begin{subfigure}[b]{0.48\columnwidth}
\centering
  \includegraphics[width=\columnwidth]{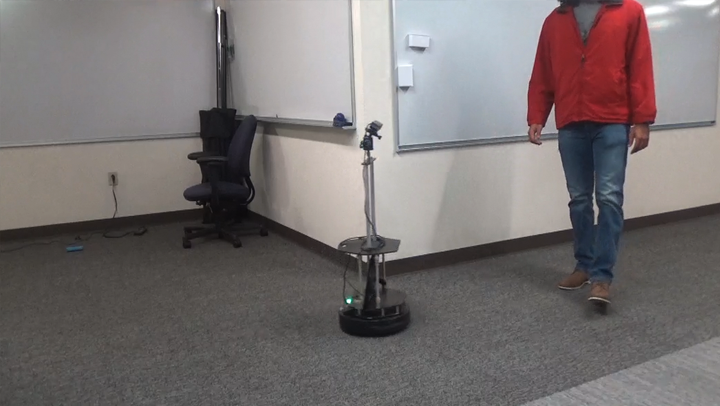}
\end{subfigure}%
\hfill
\begin{subfigure}[b]{0.48\columnwidth}
\centering

    \includegraphics[width=\columnwidth]{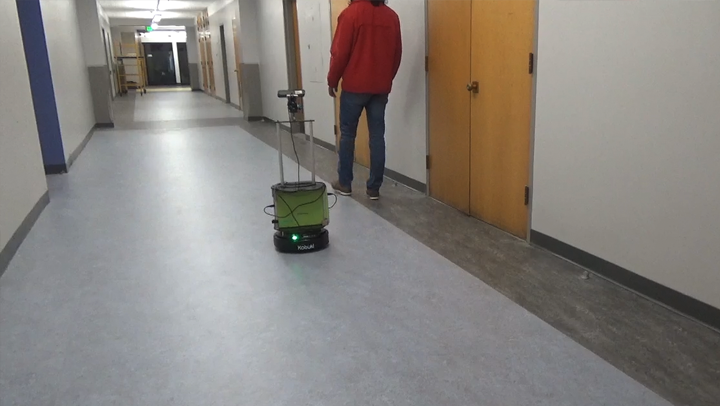}
\end{subfigure}%
\caption{Some representative images of the experiment scenarios. None of these buildings were used for training/testing purposes in simulation.}
\label{fig:experiment_scenarios}
\end{figure}

On our hardware platform, we test the robot in two buildings, neither of which is a part of \newdataset. \metName generalizes well to these new buildings and to \emph{real} humans, even though it is trained entirely on simulated data, demonstrating its sim-to-real transfer capabilities. Representative images of our experiment environments are shown in Figure \ref{fig:experiment_scenarios}.


\end{document}